%% file: main.tex
\documentclass[sigconf]{acmart}

\AtBeginDocument{%
  \providecommand\BibTeX{{%
    \normalfont B\kern-0.5em{\scshape i\kern-0.25em b}\kern-0.8em\TeX}}}

\copyrightyear{2024}
\acmYear{2024}

\begin{document}

\title{CinePreGen: Camera Controllable Video Previsualization via Engine-powered Diffusion}


\author{Yiran Chen}
\email{chenyiran@pjlab.org.cn}
\affiliation{%
  \institution{HKUST(GZ)}
  \city{Guangzhou}
  \country{China}
}

\author{Anyi Rao}
\authornote{Corresponding author.}
\email{anyirao@stanford.edu}
\affiliation{%
  \institution{Stanford University}
  \city{Stanford}
  \state{CA}
  \country{USA}
}

\author{Xuekun Jiang}
\email{jiangxuekun@pjlab.org.cn}
\affiliation{%
  \institution{sh-AILab}
  \city{ShangHai}
  \country{China}
}

\author{Shishi Xiao}
\email{sxiao713@connect.hkust-gz.edu.cn}
\affiliation{%
  \institution{HKUST(GZ)}
  \city{Guangzhou}
  \country{China}
}

\author{Ruiqing Ma}
\email{alex.ma@sjtu.edu.cn}
\affiliation{%
  \institution{HKUST}
  \city{Hong Kong}
  \country{China}
}

\author{Zeyu Wang}
\email{zeyuwang@ust.hk}
\affiliation{%
  \institution{HKUST(GZ)}
  \city{Guangzhou}
  \country{China}
}

\author{Hui Xiong}
\email{xionghui@ust.hk}
\affiliation{%
  \institution{HKUST(GZ)}
  \city{Guangzhou}
  \country{China}
}

\author{Bo Dai}
\email{daibo@pjlab.org.cn}
\affiliation{%
  \institution{sh-AILab}
  \city{ShangHai}
  \country{China}
}

\newcommand{\model}{CinePreGen\@\xspace}
\newcommand{\etal}{\textit{et al}\@\xspace}

\newcommand{\colornote}[3]{{\color{#1}\bf{#2: #3}\normalfont}}
\newcommand{\xuekun}[1]{\colornote{red}{XJ}{#1}}
\newcommand{\yiran}[1]{\colornote{blue}{YR}{#1}}
\newcommand{\anyi}[1]{\colornote{red}{AR}{#1}}
\newcommand{\xss}[1]{\colornote{green}{X44}{#1}}

\input{sections/0abstract}
\maketitle
\input{sections/1intro}

\input{sections/2related}

\input{sections/3formative}

\input{sections/3-1interface}
\input{sections/4method}

\input{sections/5exp}
\input{sections/6discussion}

\input{sections/7limitations}

\input{sections/8conclusion}


\bibliographystyle{ACM-Reference-Format}
\bibliography{main}

\input{sections/9}

\end{document}

%% file: sections/0abstract.tex
\begin{abstract}

With advancements in video generative AI models (e.g., SORA), creators are increasingly using these techniques to enhance video previsualization. However, they face challenges with incomplete and mismatched AI workflows. Existing methods mainly rely on text descriptions and struggle with camera placement, a key component of previsualization. To address these issues, we introduce CinePreGen, a visual previsualization system enhanced with engine-powered diffusion. It features a novel camera and storyboard interface that offers dynamic control, from global to local camera adjustments. This is combined with a user-friendly AI rendering workflow, which aims to achieve consistent results through multi-masked IP-Adapter and engine simulation guidelines.
In our comprehensive evaluation study, we demonstrate that our system reduces development viscosity (i.e., the complexity and challenges in the development process), meets users' needs for extensive control and iteration in the design process, and outperforms other AI video production workflows in cinematic camera movement, as shown by our experiments and a within-subjects user study.
With its intuitive camera controls and realistic rendering of camera motion, CinePreGen shows great potential for improving video production for both individual creators and industry professionals.

\end{abstract}

\ccsdesc[100]{Information systems~Multimedia content creation}

\keywords{pre-visualization, video, cinematic camera control, engine-powered diffusion}

\begin{teaserfigure}
  \includegraphics[width=\textwidth]{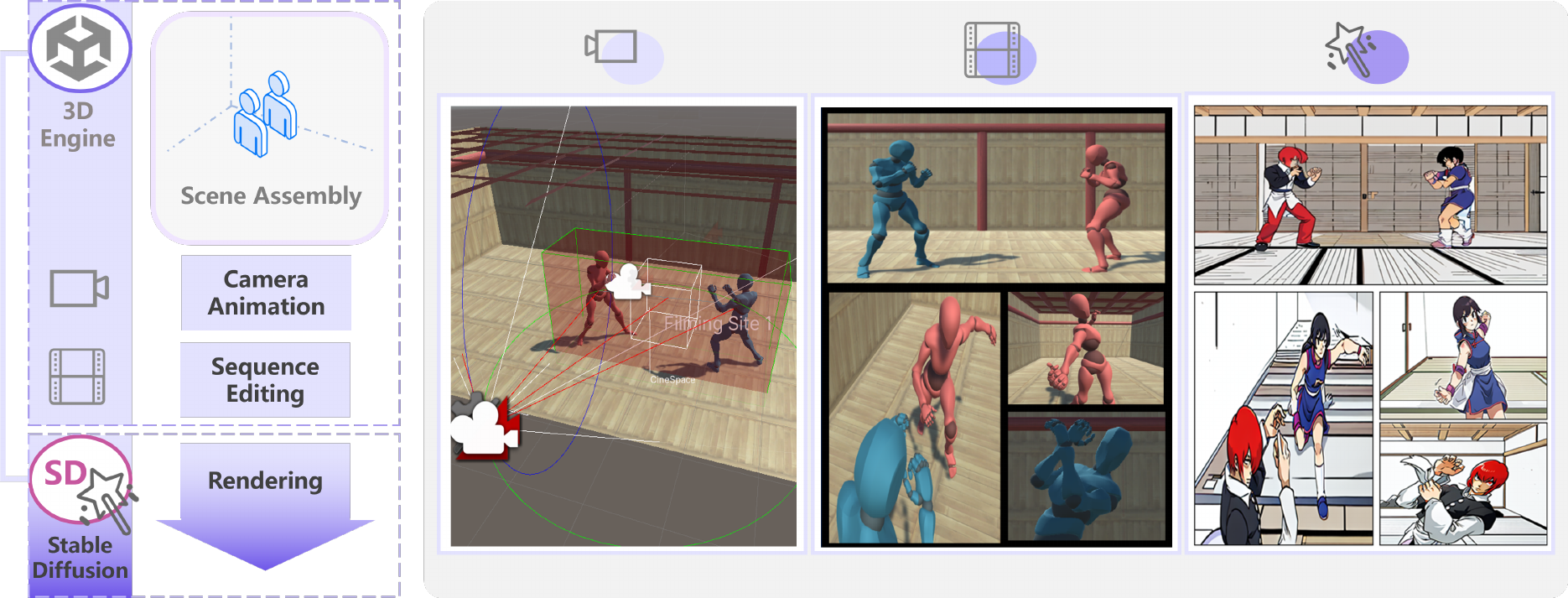}
  \caption{CinePreGen offers users flexible camera control to generate various types of cinematography, utilizing ground truth information within the engine to guide a diffusion model for consistent video generation under dynamic camera motion for video previsualization.}
  \Description{}
  \label{fig:teaser}
\end{teaserfigure}

%% file: sections/1intro.tex
\section{Introduction}
\label{sec:intro}

Can we tell a movie clip using only words? In an age where video-generative AI models are increasingly employed in previsualization for visual storytelling, we might wonder: Can we truly convey the cinematic experience through text alone? While these models allow for quick and diverse visual effects that would otherwise be labor-intensive and costly in traditional workflows, they still struggle with capturing the subtle aspects of cinematography, particularly in terms of camera angles and movements.

Traditional filmmaking relies on meticulous storyboarding and cinematography planning to guide the viewer's focus and set the emotional tone of a scene. In contrast, current AI tools in storytelling visualization predominantly focus on generating content within the frame, often neglecting the crucial aspect of camera dynamics. Though some approaches attempt to control motion in video generation, they typically address only object movement or offer limited camera control.

The main reason for this struggle in generating cinematic videos isthe scarcity of datasets annotated with detailed camera and object movements. Describing such intricate camera settings using text alone is naturally difficult, as words fall short of capturing precise positions, angles, or the dynamics of movement.

Therefore, offering camera planning for users in AI-generated videos faces challenges that extend beyond just enabling controllable adjustments; it also involves the fundamental issue of the base model's lack of training in camera motion. Considering the existing challenges, achieving professional-level film and video generation remains a distant goal. Thus, our research is focused on achieving a certain level of visual effect with highly controllable camera movements, which meet previsualization needs.

In this paper, we introduce CinePreGen, a video previsualization system that combines game engines and diffusion models. Users can efficiently achieve cinematic camera movements like pan, tilt, push, Dolly Zoom, and trucking, while also creating custom movements and compositions within our engine's camera planning system.

To enhance flexible camera adjustments and shot creation, we introduce CineSpace, an efficient representation for camera parameter space that incorporates toric space which benefits the design of two-shot. With CineSpace, the camera extrinsics parameters can easily mapped to meaningful camera behavior, and then achieve complex camera movements.
For improved storytelling, we utilize a hierarchical approach that integrates storyboards with customizable camera behaviors and precise poses. It automates camera pose calculations through a blend of predefined movements and user-defined parameters, using Bezier curves for tracking shots and CineSpace for smooth interpolations. This setup enables creators to swiftly prototype and refine their visions, enhancing narrative flow and visual storytelling efficiency.

In our rendering system, we tackle the challenge of aligning advanced cinematic camera movements with video generation from simplified models in the engine. We enhance conditional control for video generation by exporting ground truth data, such as depth and motion from the 3D environment, improving accuracy over direct video estimations and enabling precise manipulation of scene and character visuals. Through the strategic use of masks and advanced personalized video models, we ensure style and identity consistency across scenes and characters, facilitating targeted generation while preserving important cinematic camera movements.

We conducted a within-subjects user study to assess the usability and performance of our system. Our primary focus was on evaluating the cinematographic quality of the generated content and gathering user feedback regarding the manipulation of camera angles within a 3D space. The findings suggest that our interface is characterized by enhanced user-friendliness and adept at achieving professional cinematographic effects with coherent rendering and realistic camera movement.

To summarize, this paper offers the following contributions:

\textbf{(1) CinePreGen System}: A visual previsualization system that merges game engines with diffusion models, enabling efficient camera control and coherent rendering to create videos with cinematic camera movements.


\textbf{(2) Evaluation}: We conducted a within-subjects user study with 12 participants to evaluate the usability and effectiveness of CinePreGen in enhancing camera movement in visual previsualization. 

\textbf{(3) Discussion}: We highlight the importance of personal creativity in cinematic shooting, the innovative future perspective of camera movement collages for unique content generation, advocating for tools that reflect individual creativity and foster a deeper sense of user ownership, engagement, and satisfaction in the video production process.



%% file: sections/2related.tex
\section{Related Work}

\subsection{Video Generation with Motion Control}
In the field of video generation, significant focus was placed on controlling object motion to better match user preferences. Extensive research efforts explored various methods to manipulate the movements of objects. Techniques such as using bounding boxes for user-driven control of motion~\cite{chen2024motionzero,yang2024directavideo} and altering object trajectories~\cite{wang2023motionctrl,jain2023peekaboo} were developed. VideoComposer~\cite{wang2024videocomposer} enhanced this control by offering global motion guidance through pixel-wise motion vectors, and other video editing tools~\cite{bai2024uniedit,qi2023fatezero} facilitated motion edits using text-driven or manual specifications, demanding consistency between frames. Collectively, these methods primarily targeted object motion control, focusing on high semantic levels within localized scenes.

Conversely, camera motion control received less attention. Solutions like AnimateDiff~\cite{guo2023animatediff} utilized temporal LoRA modules~\cite{hu2021lora} trained on specific video sets, restricting the scope of camera movement control to types represented in the training data. MotionCtrl~\cite{wang2023motionctrl} built a video dataset annotated with camera poses to study camera motions, a process that required extensive manual input. Direct-a-video~\cite{yang2024directavideo} attempted to simplify this by adding predefined camera movements to existing videos, which could minimize the need for extensive annotations but still depended heavily on fine-tuning pre-trained video generation models. This reliance on intensive computation and model adjustments limited the variety of camera motions and generally confined them to predefined styles found in training datasets. Nevertheless, these methods generally lacked the precision needed to distinctly separate camera and object motions within videos. The challenge remained to develop more refined and flexible techniques that could emulate the dynamic and cinematic quality of camera work without the extensive computational overhead and reliance on pre-existing model constraints.

\subsection{Storyboard and PreVis tools}
Storyboards are investigated in keyframe or text summarization from videos~\cite{mohanta2013novel, bhaumik2015real, ronfard2022prose}, textual script writing~\cite{chandu2019storyboarding,mirowski2022co} and video creation assistance~\cite{goldman2006schematic,ye2008towards,pizzi2010automatic}.
Among the most relevant ones,
\citet{ye2008towards} focus on the language mapping between the action descriptions and avatar animation instead of the visual content, and \citet{pizzi2010automatic} only generate sketch-style static images.
Besides, intelligent creation tools are 
in great demand as they can help users efficiently create customized dynamic content, e.g. video, animation~\cite{louarn2018automated,louarn2020interactive}.
Cine-AI~\cite{evin2022cine} is a procedural cinematography toolset capable of generating in-game cut scenes in the style of eminent human directors.
Virtual Dynamic Storyboard~\cite{rao2023dynamic} proposes a novel virtual dynamic storyboard approach in an efficient and customized way for amateurs and builds a sequential storyboard from a script.
Some researchers focus on several key steps, such as frame composition~\cite{zhong2021aesthetic}, shot selection~\cite{liao2020occlusion}, shot cut suggestion~\cite{pardo2021learning}.
Others tackle high-level automatic procedures with simple user interactions and take multiple videos captured by different cameras to produce a coherent video in different application scenarios~\cite{arev2014automatic,leake2017computational,truong2019tool} 
using different data sources~\cite{chi2021automatic, moorthy2020gazed, rao2022shoot360}.
Different from previous methods, our system is a holistic content creation framework, which not only provides a convenient and simple camera design tool but also provides an AI rendering pipeline.

\subsection{AI Storytelling Visualization}
With the development of generative models in the fields of image generation and video generation, more and more researchers have begun to explore its potential application in storytelling visualization.
Most research is divided into two forms: image Illustrating and dynamic videos.
Avrahami~\textit{et al.}~\cite{avrahami2023chosen} proposed a method to obtain character characteristics by continuous clustering, iterative condensation step by step, and finally generating consistent characters.
Pan~\textit{et al.}~\cite{pan2024synthesizing} proposed a latent diffusion model auto-regressively conditioned on history captions and generated images, leveraging diffusion models for coherent visual story synthesis.
Liu~\textit{et al.}~\cite{liu2023intelligent} proposed a learning-based auto-regressive image generation model, termed as StoryGen.
AutoStory~\cite{wang2023autostory} aimed to generate a series of images that match the story described in texts with minimal human interactions. The user needs to specify the input story script and the characters in the image.
Similar to AutoStory, TaleCrafter~\textit{et al.}~\cite{gong2023talecrafter} proposed a versatile and generic story visualization system that leverages large language and pre-trained T2I models for generating a video from a story in plain text.
Other studies are looking to directly generate a video to improve the expressiveness of stories~\cite{yin2023nuwa, wang2024aesopagent}.
To better control the video generated, AnimateAStory~\textit{et al.}~\cite{he2023animate} presented a novel retrieval-augmented paradigm for storytelling video synthesis. The narrative elements carried in the reference video are used to constrain the generated video.

The existing methods, whether for image Illustrating or dynamic videos, are all focused on how to generate consistent characters in different images or continuous videos. They all ignore the importance of cinematography in the storytelling.
Unlike previous approaches, our approach not only solves the problem of character consistency, but also uses the game engine to bring control over cinematography into storytelling visualization to help creators create more engaging content.




%% file: sections/3formative.tex
\section{Formative Study}
\begin{figure*}[h]
    \centering
    \includegraphics[width=\linewidth]{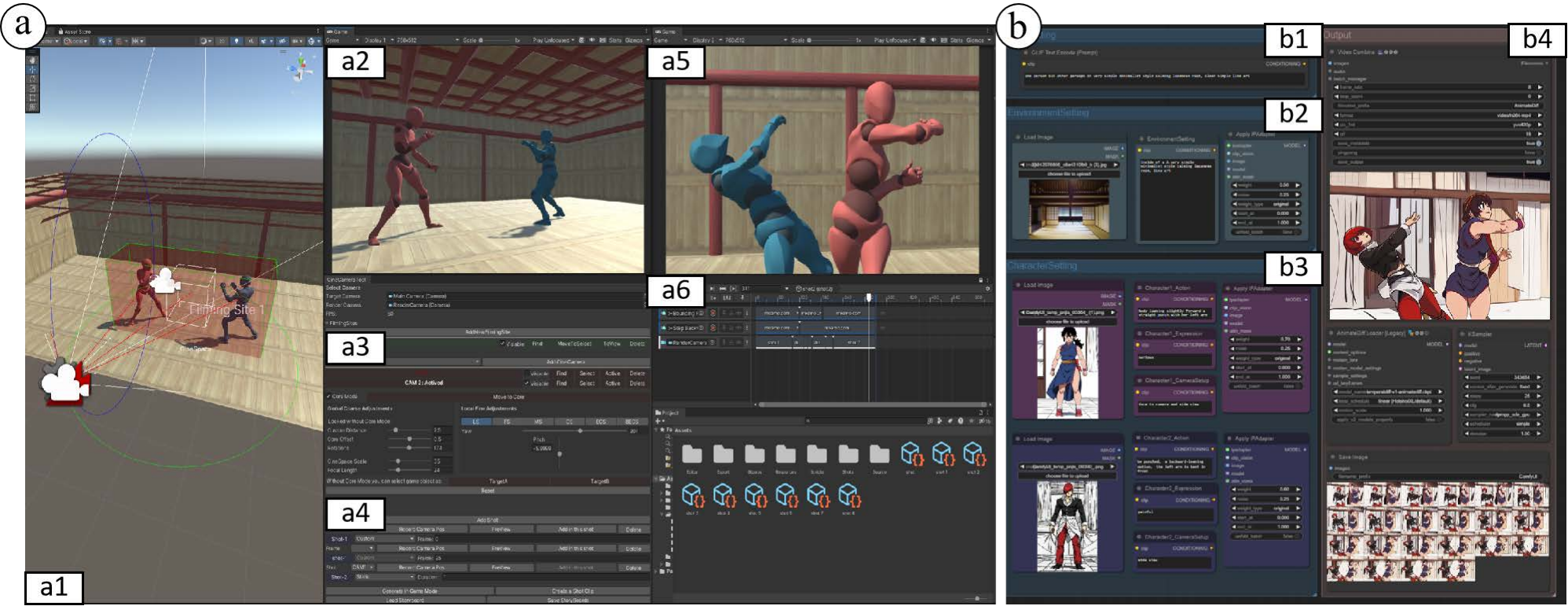}
    \caption{CinePreGen consists of two sub-modules: the layout design module (a) and the AI rendering module (b). (a1) The 3D viewer of the entire scene. (a2) The current active camera viewer. (a3) The camera editor. (a4) The storyboard editor. (a5) The preview viewer. (a6) The timeline. (b1) The shot prompts panel. (b2) The environment settings panel. (b3) The character settings panel. (b4) The output preview panel.}
    \label{fig:interface}
\end{figure*}

To understand the role and challenges of AI in visual previsualization, we conducted interviews with industry professionals who have extensive experience in both traditional film production and AI previsualization tools. We spoke with three visual content creators, each having three years of experience in traditional filmmaking and one year of intensive use of AI tools. This combination of expertise allowed us to gain insights into both the traditional workflow and the emerging AI-assisted processes. The semi-structured interviews, lasting 30 to 60 minutes, were conducted in person and focused on three key areas: 1) the use of generative AI in video creation; 2) challenges encountered with AI technology; and 3) strategies to overcome these challenges.
\subsection{Traditional Previsualization}
From these discussions, it became clear that traditional previsualization relies heavily on storyboarding and, in some cases, advanced 3D previews to plan scenes and camera movements. For early scene creation and shot planning, creators often use 3D software, supplementing their work with assets from 3D asset websites. However, arranging complex camera movements within these traditional tools is time-consuming and can be a significant bottleneck in the production process. Storyboards serve as a foundational tool for organizing shooting schedules, ensuring effective storytelling, and managing production costs.
\subsection{AI Pipeline Challenge}
The interviews revealed that professionals are integrating generative AI models into their workflows primarily for tasks like character design, concept art, and mood board creation. While these tools are valuable for generating preliminary visual ideas, they fall short in handling more intricate aspects of film production, such as camera movement and shot coherence. One professional described using AI tools like Midjourney for storyboarding but noted difficulties in translating these AI-generated visuals into coherent videos, particularly when both characters and the camera are in motion. The resulting videos often exhibited inconsistencies, such as the generation of unrelated objects in scenes that required precise focus on characters. This underscores a key issue: while AI tools can generate visually compelling content, they struggle with maintaining visual and narrative coherence, especially in dynamic scenes with complex camera movements.

Based on insights from these interviews, we defined a controllable AI previsualization pipeline which comprises six stages: (1) Idea Development, where the creator formulates a preliminary concept in text or sketches; (2) Model Selection, involving the choice of models trained on different data to create varied styles; (3) Layout Design, which aligns the result with the initial idea using methods like text descriptions, rectangular boxes, or reference images/videos; (4) Constraint Guidance, where additional neural network structures like ControlNet refine results by using extracted features from reference images; (5) Prompt Design, crucial for influencing the generated outcomes through various combinations of prompts; and (6) Generation, the final step where users generate and iterate through multiple outputs to achieve the desired result.

\subsection{Design Consideration}

Based on the existing AI previsualization pipeline and considering real production workflows, we aim to integrate our system with traditional pipelines, thereby enhancing the overall workflow. We have identified two key stages to improve the AI creative experience: layout design and constraint guidance. We have established two design objectives to guide the development of CinePreGen, ensuring a more effective creation experience in these stages.

\textbf{Flexible Camera Adjustment and Shot Creation:} We aim to give users extensive control over camera positioning and shot effects, backed by robust preview and editing tools. This flexibility is crucial for adjusting visual perspectives and enhancing storytelling before finalizing projects. Such features enable users to achieve the visual effects and narratives they desire in their work.

\textbf{User-friendly AI Rendering Workflow:} The current process of AI generation is often cumbersome, requiring users to navigate between multiple operation panels, resulting in a diversion of focus from the actual content creation. Therefore, a user-friendly AI Rendering Workflow holds significant potential to enhance creators' productivity by alleviating such challenges.

%% file: sections/3-1interface.tex
\section{User Interface}
\label{sec:interface}

Our system consist of two sub-modules, layout design module Fig.~\ref{fig:interface}-a and AI rendering module Fig. ~\ref{fig:interface}-b. The user designed layout data can be passed directly to the AI rendering module for generating results. Users can switch between the two modules at any time and control the generated results by adjusting the layout.

Our layout design module expands on UI elements native to traditional 3D game engine, like unity~\cite{unity}. Users can arbitrarily adjust the transform of objects (cameras or characters) in the 3d scene and preview the state of the entire scene (Fig.~\ref{fig:interface}-a1). Users can create multiple cameras in a scene and view the current active cameras (Fig.~\ref{fig:interface}-a2). The options for interacting with cameras and storyboards are located at the bottom of the active camera viewer. Under the Edit menu, the user can adjust the camera's attitude arbitrarily through the slider (Fig.~\ref{fig:interface}-a3).Once each camera state is set, user can design each shot in the storyboard menu (Fig.~\ref{fig:interface}-a4). By clicking the Generate button, user can save the shot information as an asset, and edit and preview it on the timeline (Fig.~\ref{fig:interface}-a5 and Fig.~\ref{fig:interface}-a6).

We design our rendering interface based on ComfyUI~\cite{ComfyUI}, a node-based graphical user interface for Stable Diffusion, which allows us to construct and execute customized pipelines and makes it user-friendly. Users can first input the overall design of the shot, followed by settings for scenes and characters and corresponding referenced image (Fig.~\ref{fig:interface}-b1 and Fig.~\ref{fig:interface}-b2 and Fig.~\ref{fig:interface}-b3) . This approach allows for precise control and generation of each object through separate prompts and features of reference images. The generated videos are then conveniently available for preview on the right(Fig.~\ref{fig:interface}-b4).

%% file: sections/4method.tex
\section{Implementation}
\label{sec:emethod}
\subsection{Cinematic Camera Space}
\label{sec:cinespace}

\begin{figure}[tp]
    \centering
    \includegraphics[width=\linewidth]{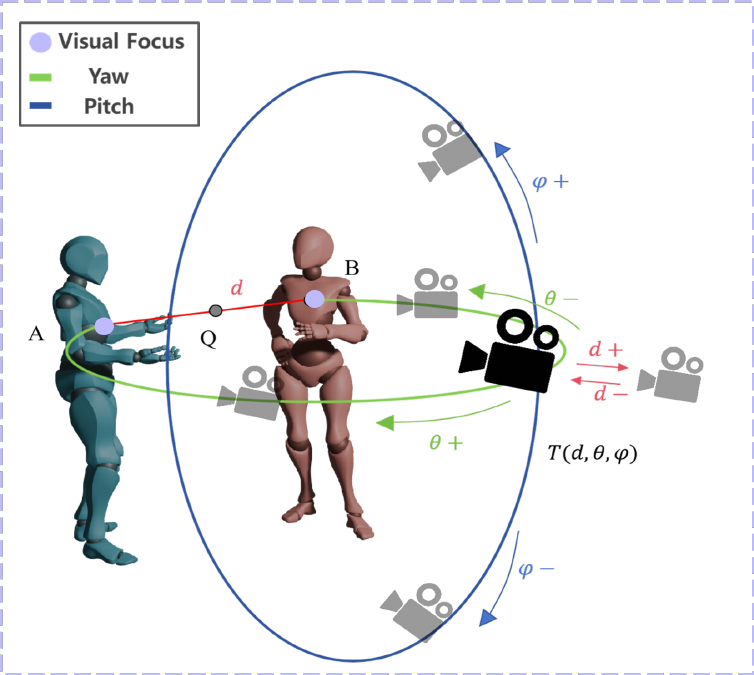}
    \vspace{-20pt}
    \caption{CineSpace, a novel and efficient representation for camera parameter space, which defined the camera's behavior by three key parameters $(d,\theta, \varphi)$.
    }
    \label{fig:cinespace}
    \vspace{-1pt}
\end{figure}

To provide dynamic control of camera placement, we introduce a novel and efficient representation for camera parameter space, namely CineSpace, as shown in Fig.~\ref{fig:cinespace}.
With CineSpace, the camera extrinsic parameters can be easily mapped to meaningful camera behavior, allowing for complex camera movements.

In the previous preview tool \cite{rao2023dynamic}, the most commonly used camera space is the spherical coordinate system, which uses radial distance $r$, polar angle $\theta$, and azimuthal angle $\varphi$ to facilitate control. The disadvantage of the spherical space is that it is single-object centered, making it unsuitable for communication scenarios, which play an important role in storytelling. 

\begin{figure}[tp]
    \centering
    \includegraphics[width=\linewidth]{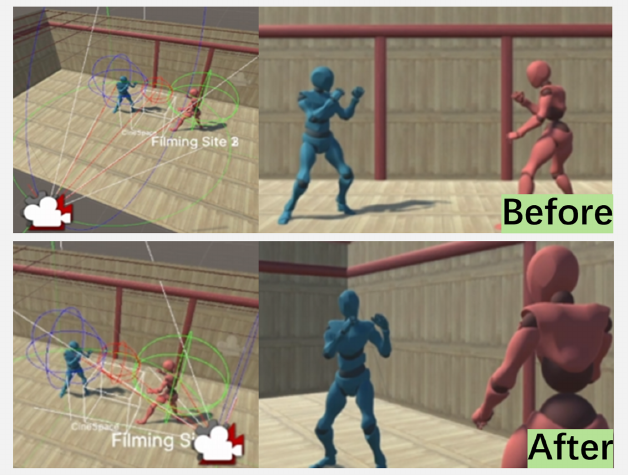}
    \vspace{-20pt}
    \caption{An example of yaw rotation in CinePreGen, where the camera uses two characters as the visual focus. As the camera yaws, it rotates based on the CineSpace coordinate system, ensuring that both characters remain in the frame throughout the movement.
    }
    \label{fig:yaw}
    \vspace{-1pt}
\end{figure}

Toric Space \cite{lino2012efficient, lino2015intuitive} is a compact representation for camera control, which represents the whole set of manifolds that may be generated around a pair of targets A and B. Toric space has a very important property: no matter how the parameters and target positions change, the two subjects (A \& B) are always in the picture, as shown in Fig.~\ref{fig:yaw}. Therefore, toric space is designed for two-shot scenarios, meaning both characters are visible together in the frame, especially in over-the-shoulder shots.
Even so, toric space has a few drawbacks. First, toric space is very sensitive to the movement of its characters. Second, it is difficult to achieve a stable PUSH IN/OUT shot. 
CineSpace combines the strengths of both camera representations while overcoming their shortcomings.

As shown in Fig.~\ref{fig:cinespace}, in CineSpace, the camera's behavior is defined by three key parameters $(d,\theta, \varphi)$. $d$ is the distance between control points $A \& B$, which means the distance between the camera and the CineSpace center $Q$, $\theta$ defines an horizontal angle around the targets and $\varphi$ defines a vertical angle around the targets. 

When the user creates a new camera in the scene, the camera is automatically associated with a CineSpace. Users can freely control the camera in the interactive interface.
To achieve more refined camera control, in addition to using CineSpace's three key parameters $(d, \theta, \varphi)$ to control camera movement on a manifold, we have also added some additional properties to CineSpace. 

\begin{figure*}[h]
    \centering
    \includegraphics[width=\linewidth]{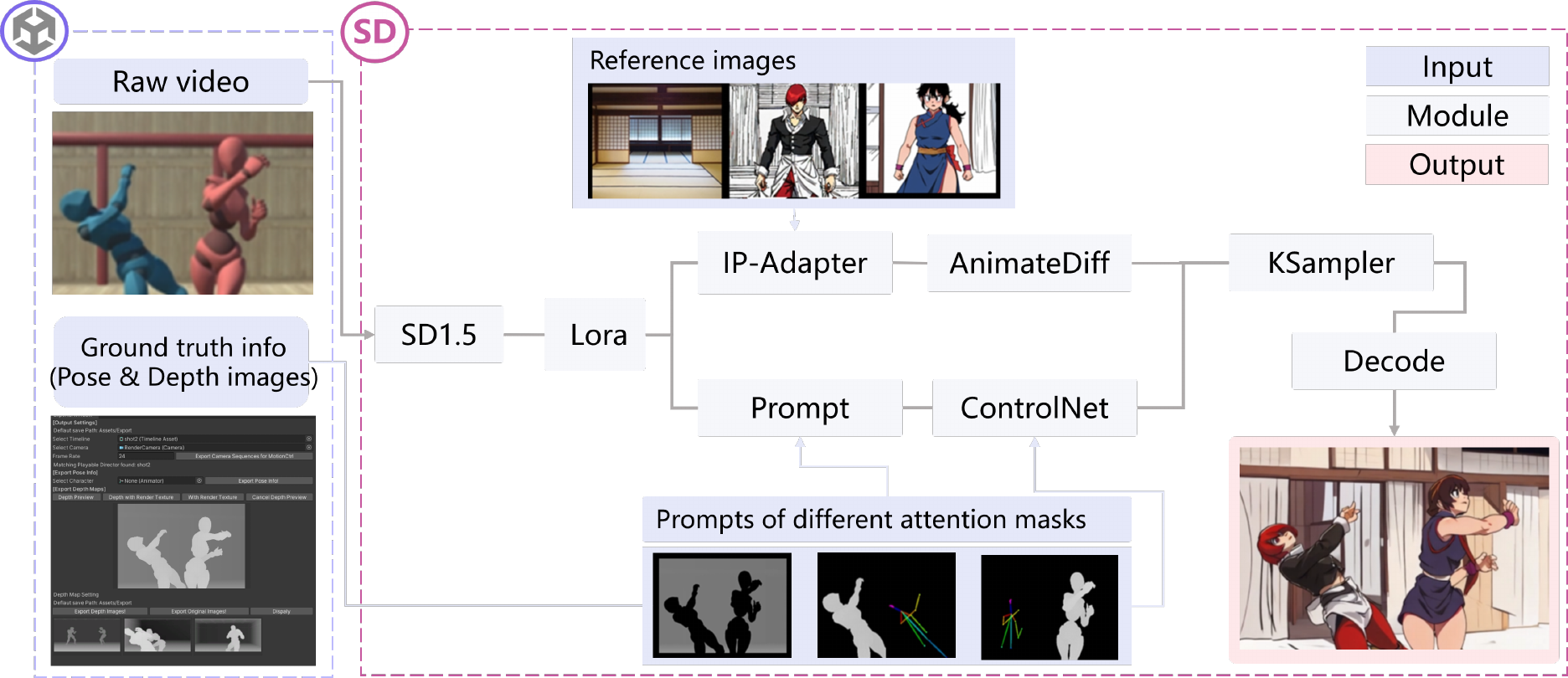}
    \caption{Diffusion Rendering Workflow—The process starts with obtaining raw footage from the engine, followed by exporting ground truth data (e.g., depth maps and pose images), applying masks for targeted control, and ensuring consistency in visual style and character identity using AnimateDiff and IP-Adapter.}
    \label{fig:ren}
\end{figure*}

For example, by controlling the proportion of the space center point from control points A and B, users can adjust the left-right offset of the screen or change the orientation of the entire CineSpace by rotating the space center point. 
Among the 7 DoF (Degrees of Freedom) of the camera, focal length is an important parameter, so we also provide focal length control in the interactive interface.
Additionally, we have defined a series of preset types based on cinematography theory (e.g., Long Shot, Close-up Shot, or Medium Shot) to help users quickly set the camera to the corresponding position. 
With these parameters, the camera can be quickly placed anywhere around the target.

\subsection{Storyboarding}

To achieve more flexible storytelling, we create a hierarchy consisting of storyboards, camera behaviors, and camera poses. 
A storyboard represents an individual shot and contains one or more sets of camera behaviors. Camera behavior refers to a meaningful camera movement (e.g., PUSH IN, PAN, TRUCKING) or an interpolation process between two camera poses. A camera pose is the state of the camera at a specific time step. At runtime, the camera pose is calculated from a set of CineSpace parameters. 
With this structure, users can control the narrative rhythm through storyboards and fine-tune the camera pose for each frame, allowing creators to quickly validate their ideas using our previsualization tool and improve their creative efficiency.

To create a new storyboard, the user needs to bind the storyboard to a camera in the scene. For each storyboard, the user can add one or more camera behaviors, and the system automatically calculates the camera pose based on the corresponding camera behaviors. 
We provide two calculation modes for camera pose: \textbf{shot mode} and \textbf{frame mode}. In our system, we implemented 15 camera behaviors commonly used in cinematography, such as PUSH IN, PAN, or TRUCKING. 

In \textbf{shot mode}, the user only needs to select the corresponding camera behavior and set the associated parameters to preview the rendering result. For example, in a PUSH IN shot, the user first sets the duration of the camera behavior, then selects the push forward range from the drop-down menu. The system then uses the camera associated with the current storyboard as the initial state and calculates the camera pose at each moment through CineSpace, based on the duration and push forward distance. %
The Tracking shot requires some additional explanation. A Tracking shot is any shot that physically moves the camera through the scene, which means that the camera must move along a trajectory. 
In our system, we use a Bezier curve to represent the camera's trajectory. In the Tracking shot drop-down menu, the user can create a new track and adjust the shape of the curve using control points. The movement of the camera is achieved by binding the trajectory to the corresponding camera behavior.

In \textbf{frame mode}, users need to associate a set of CineSpace parameters with each keyframe, and the system automatically interpolates between keyframes. Thanks to the smooth interpolation properties that CineSpace inherits from toric space, the camera motion is very fluid. 
Once all the storyboards and corresponding camera behaviors are set up, the corresponding camera poses can be generated by clicking the Generate button. Each camera pose is represented by a 4x4 transformation matrix. 
To avoid interfering with the original storyboard design, we use the generated camera poses to drive a separate rendering camera for image rendering.

\subsection{Diffusion Rendering}
As shown in Fig.~\ref{fig:ren}, in the engine's camera system, we can perform advanced and cinematic camera movements, but the footage captured using these methods does not align well with the video generation process. This misalignment arises because the scenes and characters are represented using simplified models from platforms like Mixamo, which lack the detail and realism needed for vid2vid transformations. On the other hand, using high-fidelity 3D models paradoxically limits the diversity and significance of video generation. 
Therefore, our objective is to use diffusion models to stably render these simplified scenes and characters while preserving the cinematic camera movements of the video.

To provide better conditional control, we leveraged the simulation features of the engine to export ground truth information from the 3D virtual environment, such as depth and motion data. By referencing the conditional input format of the diffusion model, we can export high-quality depth maps of different objects and pose images of characters. These are projected by the camera from skeleton positions and then transferred to an OpenPose template. This ground truth information significantly enhances control over video generation compared to results obtained from directly estimating depth and pose information from videos.

\begin{figure*}[h]
    \centering
    \includegraphics[width=\linewidth]{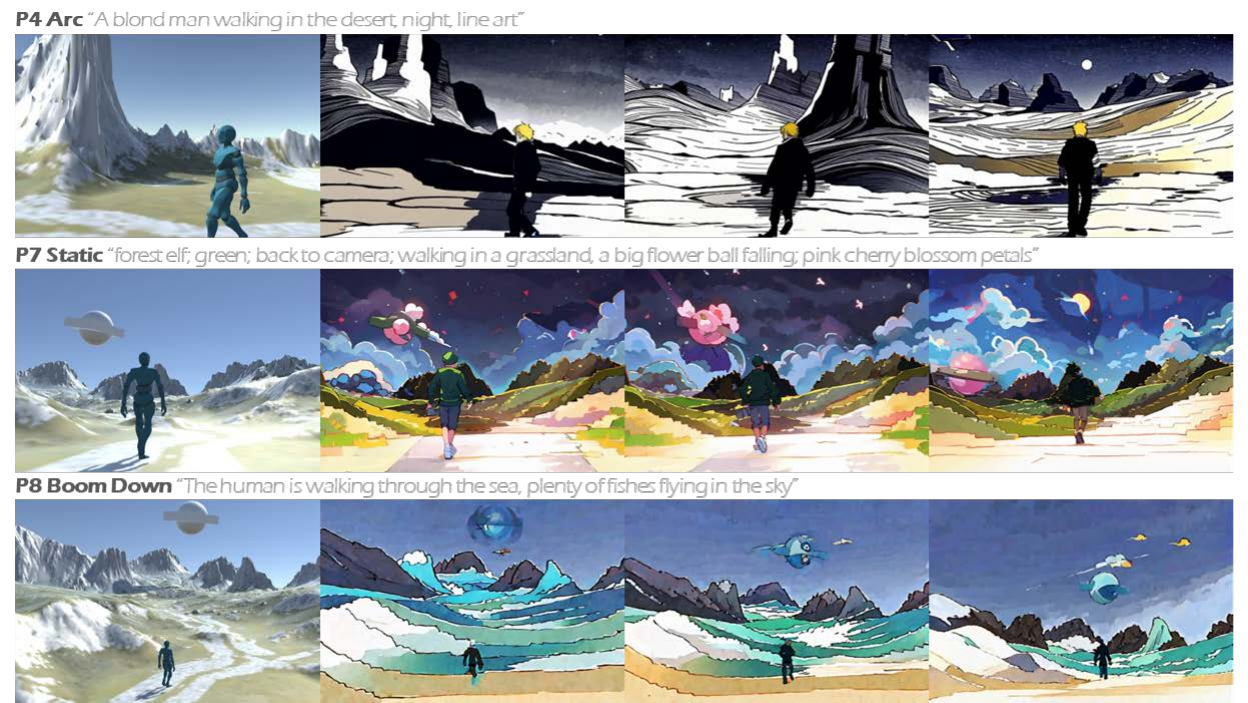}
    \caption{Participants' work by using CinePreGen, with the left side showing the original footage of camera movements generated in the engine, and the right side displaying the rendered results with the annotation of the shot type and their prompts.}
    \label{fig:user_shot}
\end{figure*}

To ensure coherence and consistency across both the scene and multiple characters, we first utilize masks selected from various rendering objects within the engine. These masks facilitate the application of different prompts to distinct areas, preventing cross-contamination among prompts and allowing for precise control over specific scenes or characters. We achieve consistency in two main aspects: visual style and character identity. With the implementation of AnimateDiff \cite{guo2023animatediff}, an animating personalized T2I model, we maintain style consistency. Furthermore, to closely align the generated content with the features of characters and scene images, we employ the IP-Adapter \cite{ye2023ip-adapter}, a text-compatible image prompt adapter. This method ensures that the generated content matches the desired features and focuses generation on specific areas requiring attention, as indicated by our provided masks. Additionally, the integration of these masks preserves camera movement information, thereby enhancing the overall visual experience.

%% file: sections/5exp.tex

\section{User Study}

We conducted a within-subjects user study with 12 participants to evaluate the usability and effectiveness of CinePreGen in enhancing camera movement in visual previsualization. We compared it with another tool aimed at customizing motion for AI animation, Deforum~\cite{deforum} which unlike CinePreGen, operates with adjustment parameters in the 2D plane. Both quantitative and qualitative results demonstrate the effectiveness of CinePreGen. 

By examining the participants' creative processes, including idea development, prompt design, and the choice of camera movements across different periods, we further validated the effectiveness of CinePreGen.

\subsection{Participants}
12 participants took part in the study, consisting of 5 females and 7 males, aged between 21 and 29 years old. Among these participants, 10 majored in animation or game design, while 2 came from the film-making industry. All participants possessed experience with camera movement and storyboarding in their creation process. Prior to the study, they had been utilizing 3D modeling software such as Unity 3D and had experimented with AI-generated content tools like MidJourney and Stable Diffusion.
As a token of appreciation, each participant received a \$20 gift card upon completing the study.


\subsection{Study Protocol}
We conducted our evaluation through a task-based usability study involving two tools for personalization camera movement in visual previsualization: CinePreGen and Deforum. Deforum is a widely used motion effect extension that can be integrated into Automatic Web UI, enabling control over camera movement by specifying transformation strength in two aspects: translation XYZ and rotation XYZ.
To mitigate bias, we randomized the order in which participants experienced these two tools. Throughout the study, we encouraged participants to think aloud, and we recorded both their voices and interaction logs to capture their habits in designing camera movements. Semi-structured interviews were also conducted to gather feedback and insights from the participants.

\subsubsection{Briefing (15 minutes).}
We firstly presented a video about different camera movement leverage in movies to introduce the topic. 
To facilitate a swift transition for participants into their roles as creators, we inquired about their impressive camera movement in movies or from their prior experience.
Next, we show several outputs of current video generation model and then asked them to think about what if they design this camera movement.

\subsubsection{Operation Demonstration (10 minutes).}
We demonstrated to users the interface of our system's camera control within the engine, showcasing the usage workflow and methods, given a two-player combat Unity project as an example.
\begin{figure}[h]
    \centering
    \includegraphics[width=\linewidth]{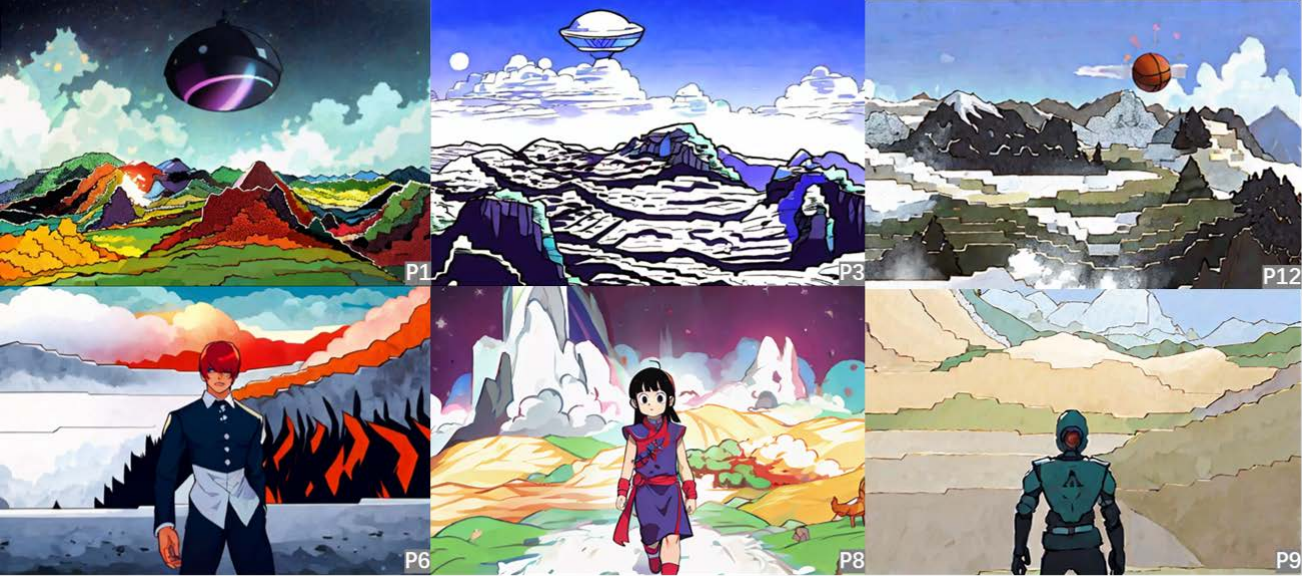}
    \caption{Different scenes and characters generate by participants in CinePreGen.}
    \label{fig:user_setting}
\end{figure}

\subsubsection{Task (35 minutes).}


To simulate the previsualization phase of video production, we tasked participants with the scenario of video generations a walking person encountering a falling object in a world. We allowed users to customize their shooting strategies and the world character styles, which correspond to camera movement and visual effect needs.

In order to ensure the comparability of results from different users, we provided the same initial Unity project, which included identical scene layouts and animations. This approach allowed us to focus on the variation in filming techniques within the same scene.

For a fair comparison between Deforum and CinePreGen, we also provided multi-angle photos from the Unity scene. These images served as initial reference inputs for Deforum, enabling users to adjust camera movements.

\begin{figure*}[h]
    \centering
    \includegraphics[width=\linewidth]{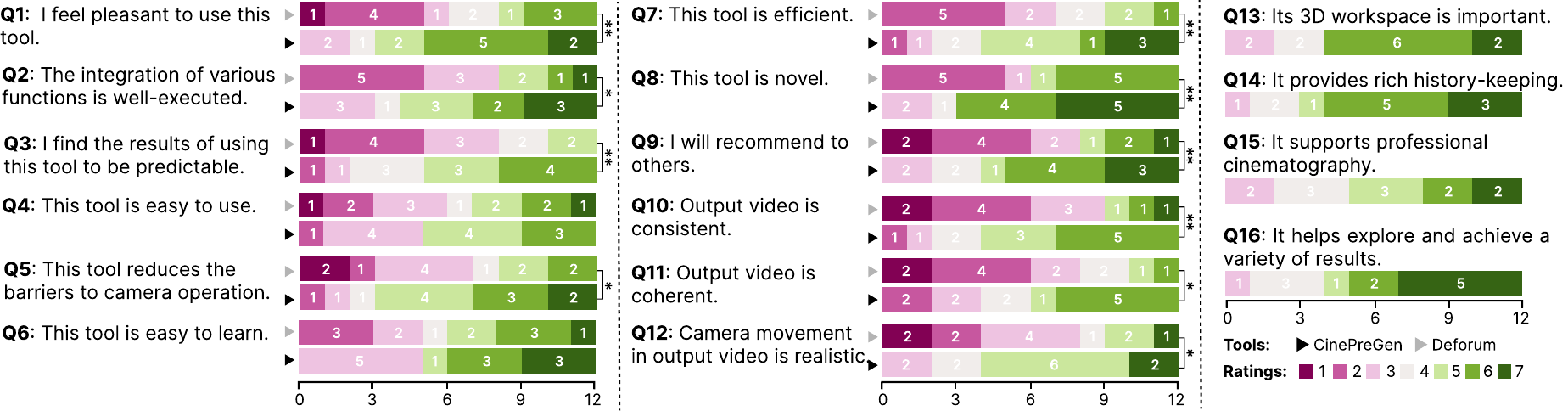}
    \caption{Participants' ratings for Deforum and CinePreGen, with * indicating \(p < 0.05\) and  ** indicating \(p < 0.01\). Besides, participants also rate the special functions of CinePreGen.}
    \label{fig:user_study}
\end{figure*}

\begin{figure}[h]
    \centering
    \includegraphics[width=\linewidth]{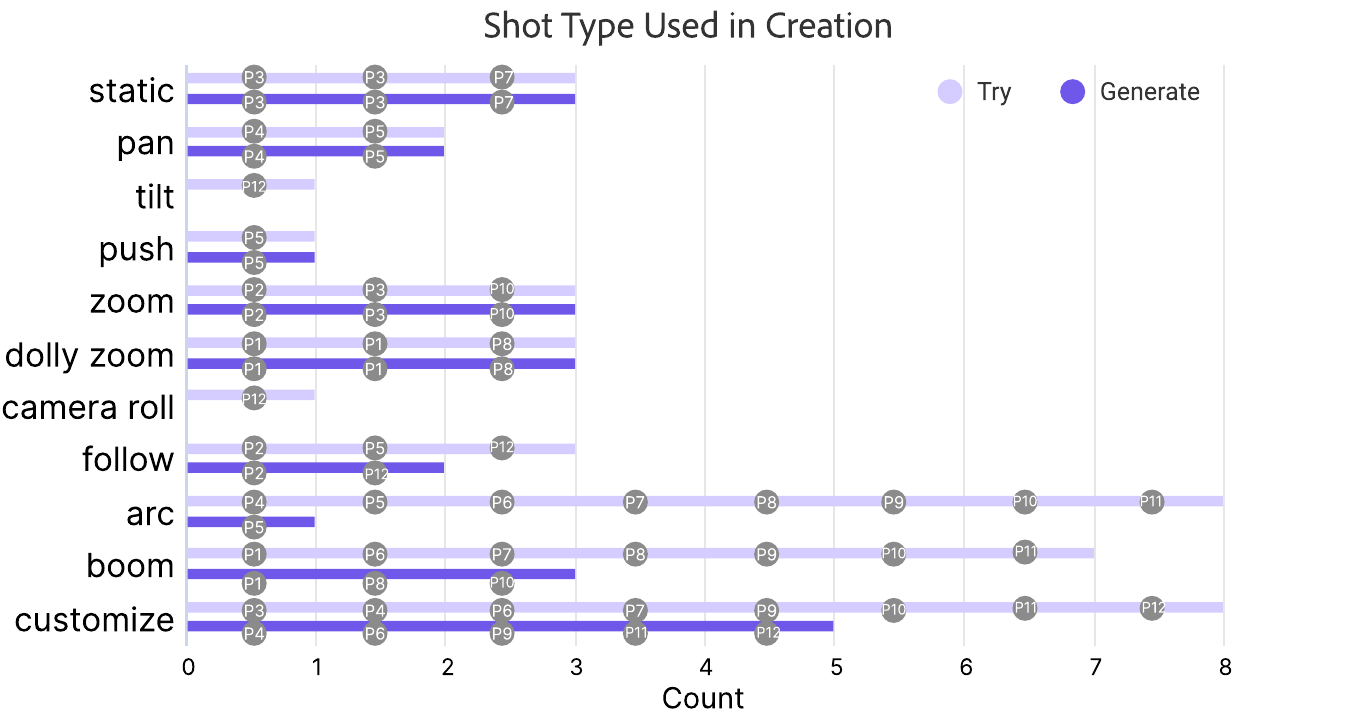}
    \caption{The shot type tested and ultimately utilized by participants in CinePreGen.}
    \label{fig:type}
\end{figure}

Considering the simplicity and comparability of the experiment, we did not include control for character consistency, which involves the IP-Adapter with multiple attention masks, user can choose lora models to generated the corresponding character. Instead, we utilized the engine's ground truth depth maps and pose maps with masks to control the rendering, thereby maintaining the desired camera movement effects.

To control the duration of the experiment, in the CinePreGen trials, users can generate multiple camera movement videos within the engine, but only 1 to 3 video clips are produced during the rendering process.

\subsubsection{Qurstionnaire and Interview (20 minutes).}
Upon completing all tasks, participants were requested to fill out a questionnaire. Since 7-point Likert scale offers a good balance by providing detailed insights without overwhelming respondents, unlike simpler 5-point scales that might lack nuance or more complex 10-point scales that could cause respondent fatigue.
We utilize it to assess the usability of both CinePreGen and Deforum. This assessment encompassed a range of aspects, including questions regarding user experience, system usability, performance of the generated video, and function evaluation.

\subsection{Analysis}
\subsubsection{Quantitative Results}
We will report the results of the quantitative questionnaire from three perspectives: 1) the user experience with the tools (Q1-Q9), 2) the performance of output video (Q10-Q12), and 3) rating for special features in CinePreGen (Q10-Q12).
Specifically, we performed the
Wilcoxon test to compare CinePreGen and the Deforum in all
aspects. 
As shown in Fig.~\ref{fig:user_study}, We denoted a statistically significant difference with “*” for \(p < 0.05\), and “**” for \(p < 0.01\).

\noindent
\textbf{User experience (Q1-Q9).}
CinePreGen outperformed the Deforum statistically significantly in terms of experience and effectiveness (Q1-Q3, Q5, Q7-Q9). Specifically, participants expressed pleasant to use CinePreGen (Q1) and willingness to recommend it (Q9) with high mean value above $5.3$, while Deforum got lower score lower than $3.6$. In terms of effectiveness, participants reported a significant difference $(p < 0.01)$. 
Aspects such as integrated function (Q2) and reduced barrier for camera movement (Q5) also demonstrated statistically significant differences $(p < 0.05)$.
However, there is no significant difference between CinePreGen and Deforum due to the professional expertise in mastering camera movement.
But CinPreGen presents a slightly higher mean value and lower variance.
For instance, in terms of easy to use (Q4), CinePreGen has $(\mu=4.33, \sigma=1.44)$, while Deforum gets $(\mu=3.91, \sigma=1.88)$.

\noindent
\textbf{Performance of output video (Q10-Q12).}
Significant differences were observed between CinePreGen and Deforum in the consistency (Q10) of output video\((p<0.01)\).
Over half of the participants reported that the characters and environment remained the same across different frames of the output video \((\mu=4.83, \sigma=1.34)\).
They also recognized the coherence of the generated video (Q11), which describes that the transition between two frames will not alter the role of visual representation \((p<0.05)\).
Regarding the realism of camera movement perception in the generated video, CinePreGen also showed a significant difference compared to Deforum \((p<0.05)\). This shows that even if we don't use the IP-Adapter, separating the prompt inputs from the masks combined with the ground truth of depth and pose guidance from the engine combined with the personalized video model gives a certain consistency. 

\noindent
\textbf{Rating special functions in CinePreGen (Q13-Q16).}
Overall, participants showed agreement regarding the functionality of CinePreGen in their creation process. Specifically, nearly half of the participants gave the highest score in recognizing that CinePreGen can aid exploration and achieve a variety of results (Q16).
Both the scores for appreciation of the 3D workspace (Q13) and history-keeping features (Q15) earned high average scores of $5.33$ and $5.58$, respectively.
As for the professional cinematography support (Q15), participants reported relatively lower scores, but with an appreciation for AIGC involvement in camera movement \((\mu=4.92, \sigma=1.37)\).

\noindent
\textbf{Participants' preferences for camera movements}
\noindent
We documented the shot type tested and ultimately utilized by participants in the  CinePreGen trail as shown in Fig.~\ref{fig:type}, indicating that every category of camera movement we designed was either experimented with or used. We observed a preference among users to explore Arc and Boom movements and customized camera techniques within the system. In the final selection of shots, there was a tendency towards customized camera movements. However, static, zoom, and boom shots were also more frequently chosen compared to other categories.

\subsubsection{Qualitative Results}
\textbf{Within the same story, but across unique worlds.} As shown in Fig.~\ref{fig:user_setting} and Fig.~\ref{fig:user_shot}, our users demonstrated remarkable diversity in their task's world settings. For example, Participant 7's shot was defined as "one person, walking in a grassland, a big flower ball falling," with the character set as "forest elf; green; back to the camera," and the falling object designed as "a big flower ball; fluffy balls; pink cherry blossom petals." In contrast, Participant 10's setting was "The human is walking through the sea," with the falling object being "plenty of fishes flying in the sky." Some participants, like P6 and P8, opted for the Lora model for character settings. Our system showcased a wealth of visual effects and demonstrated the capability to independently generate and control scenes, characters, and objects.

\textbf{Interacting with 3D scenes is important for camera adjustment.} Feedback from participants underlined the superior, intuitive, and creative control that CinePreGen provides over camera movements, especially when contrasted with Deforum, which confines movements to a 2D plane. Moreover, P1 pointed out the added benefit of operating within a 3D space: It not only aligns more closely with intuitive user interactions but also enhances the potential for re-editing and creatively utilizing the 3D assets present in the scene. This dual advantage underscores the critical role that 3D control space plays in elevating the user experience and expanding creative possibilities.

\textbf{Benefit for individual video creators.} P9, with a background in 3D animation, has observed that our system allows him to quickly carry out pre-visualization projects on his own. He also pointed out that for those users who do not demand high rendering quality (for instance, those who favor a cartoon or Blender 3 rendered 2 style), our system meets their creative requirements efficiently. Not only does it simplify the previsualization process, but it also provides individual creators with robust tools. These tools enable them to independently manage the entire creative process, from the initial design to the final rendering, enhancing their creative possibilities and productivity. 

\textbf{Authentic industrial needs.}
P6, a student specializing in film-making further highlighted the commercial and educational prospects of our work. From a commercial perspective, there's a demand in the film industry for pre-setting styles. Once a film has established several styles, subsequent shots and camera movements can be rapidly produced. The generation model can then create content according to these predefined styles, significantly facilitating the pre-setting process. On the educational front, the manipulation of camera movements can serve as a teaching tool, enabling students and beginners to grasp and apply these techniques effectively to achieve their desired preview effects.

%% file: sections/6discussion.tex

\section{Discussion}
\label{sec:discussion}
\textbf{Can Shooting Choices Reflect Our Unique "Fingerprint"?} Despite offering preset shot type options, the videos users create display remarkable variety. Even for the same arc shot, different starting positions, angles, and durations can result in unique visual experiences. This diversity underscores the limitations of textual descriptions, as each individual's perspective and observational style contain personalized elements that transcend words. Some studies have analyzed the combination of shot scale and type used by different directors, considering it as their unique shooting fingerprint. In AI-assisted video creation, it's crucial to provide tools that support personalized expression, allowing generated videos to reflect not only required scenes but also the creator's unique perspective and style.

\vspace{4pt} \noindent\textbf{New Collage of Camera Movement.} Our framework utilizes ground truth data from gaming engines, containing depth and pose information influenced by various cinematographic movements. We found that this data can be mixed to create new visual content. For instance, by combining export data from different times in the engine, we can generate new videos. This approach allows replication of objects within the same scene and blending of multiple objects into a single frame, each with distinct camera movements, introducing unique effects not achievable with traditional methods. This technology could become a new category in video production, adding flexibility and diversity to visual storytelling.

\vspace{4pt} \noindent\textbf{Beyond Controllability.} While our system was designed for camera control, we discovered benefits beyond mere controllability. One user noted that seeing her designed camera movements in the generated videos gave her a sense of participation and achievement, different from previous AI workflows. This suggests that AI tools must preserve user engagement and accomplishment, as these elements enhance the creative experience and sense of ownership. For some tasks, users may seek streamlined processes, while for others, richer, more immersive tools that foster creativity are more appealing.


%% file: sections/7limitations.tex
\section{Limitations and Future Work}

While CinePreGen provides an effective visual previsualization system, several limitations remain to be addressed in future work.

\textbf{Experiment Limitations.} The user study involved creating a limited set of independent shots (3-5) based on a simple plot description. This setup may not fully capture the narrative depth and influence of initial shots on subsequent ones. Future studies could benefit from randomized, controlled, and longitudinal trials to better understand CinePreGen's integration into AI creative workflows across different scenarios.

\textbf{Cinematography Knowledge.} CinePreGen assumes some user knowledge of cinematography, which may limit accessibility. To address this, we are considering adding integrated tutorials or AI-based tips for camera setups, making the platform more user-friendly for beginners while also serving as an educational tool.

\textbf{Smart Camera Positioning and Scene Adaptation.} Setting initial camera positions in large-scale scenes, such as urban environments, can be challenging. We propose using visual understanding technology to recommend suitable camera positions and dynamically adjust settings based on scene elements, enhancing both aesthetics and narrative impact in real-time.

\textbf{Collaborative Previsualization.} Expanding CinePreGen to support collaborative previsualization would benefit larger teams by enabling real-time feedback, shared editing sessions, and version control, thus improving communication and alignment in complex projects.

%% file: sections/8conclusion.tex
\section{Conclusion} 
\label{sec:conclusion}
In this work, we introduced CinePreGen, a novel approach to visual previsualization that leverages engine-powered diffusion to enhance control over camera dynamics. Our user study, involving participants from animation, game design, and filmmaking, demonstrated the system's usability and effectiveness in improving the previsualization process. Both quantitative and qualitative assessments highlighted CinePreGen's ability to provide professional camera control with history-keeping capabilities and achieve coherent rendering guided by ground truth information. By integrating engine capabilities with generative AI models, we believe our system offers significant benefits to both individual creators and industry professionals, helping them streamline workflows and enhance creativity.

We are encouraged by the potential of CinePreGen to revolutionize the previsualization landscape, providing creators with unprecedented levels of flexibility and precision. As we continue to develop this system, we anticipate its broader adoption across various creative industries, bridging the gap between traditional filmmaking techniques and modern AI advancements.

Looking forward, we see significant opportunities to further refine and expand CinePreGen's capabilities. Our future work could involve developing more intuitive interaction models, implementing adaptive AI systems that learn from user interactions, and establishing benchmarks along with creating high-quality multimodal datasets. Additionally, we plan to extend the system's capabilities to include features such as background music generation and diverse character styles, offering a more comprehensive multimedia experience.

Through this research, we aim to advance digital cinematography by integrating cutting-edge AI technologies, offering practical tools that streamline the previsualization process, and making these advanced technologies accessible to a broader range of creators.


%% file: sections/9.tex
\appendix
\renewcommand{\thefigure}{A\arabic{figure}}
\setcounter{figure}{0}

\section{Supplementary Material}

\begin{figure*}[h]
    \centering
    \includegraphics[width=\linewidth]{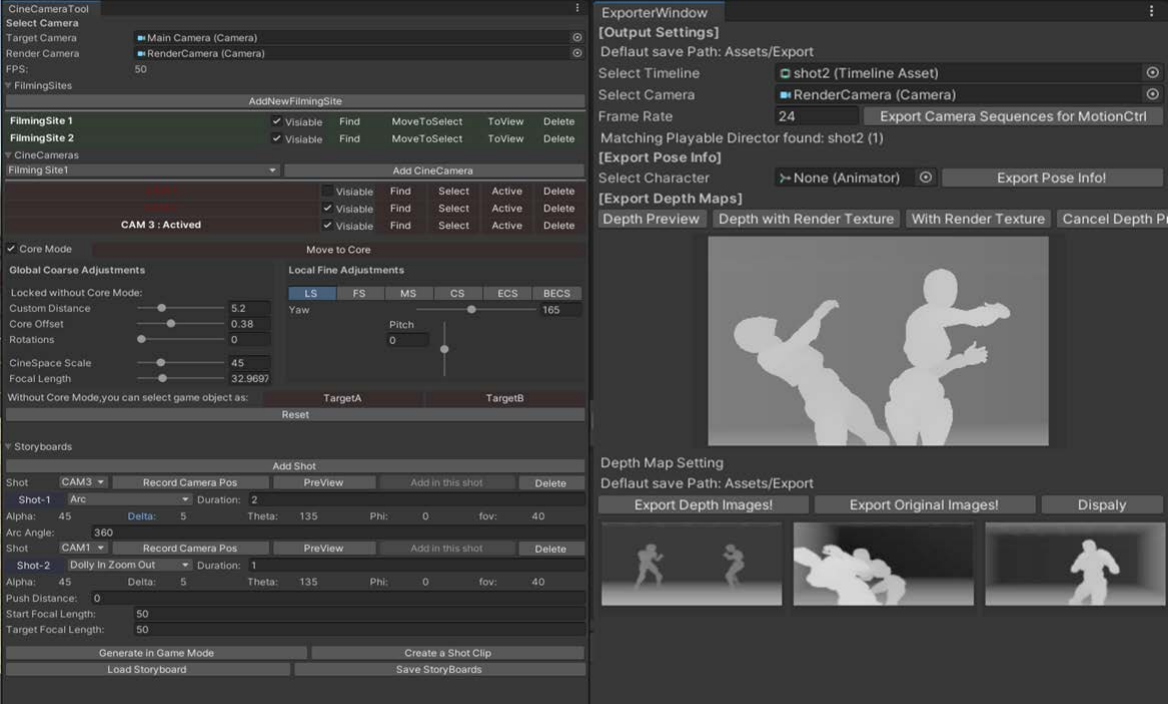}
    \caption{User interface for camera control and ground truth information export in the engine.}
    \label{fig:unity_ui}
\end{figure*}

\begin{figure*}[h]
    \centering
    \includegraphics[width=\linewidth]{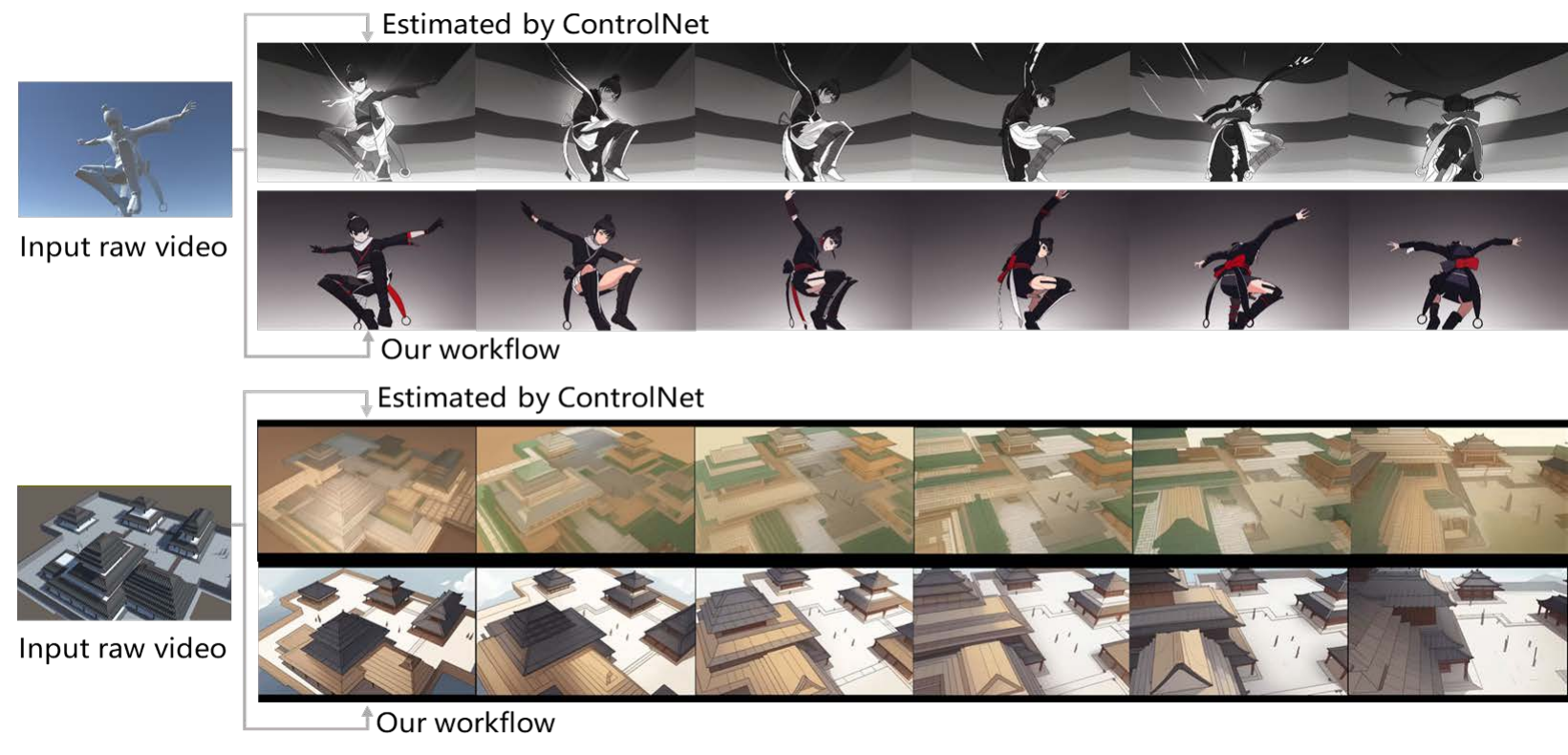}
    \caption{Comparison of generation consistency using estimated depth and pose versus ground truth information exported from the engine.}
    \label{fig:compar}
\end{figure*}